\documentclass{article}
\usepackage{ijcai22}

\usepackage{times}
\usepackage{soul}
\usepackage{url}
\usepackage{hyperref}
\usepackage[utf8]{inputenc}
\usepackage[small]{caption}
\usepackage{graphicx}
\usepackage{amsmath}
\usepackage{amsthm}
\usepackage{amsfonts}
\usepackage{booktabs}
\usepackage{algorithm}
\usepackage{algorithmic}
\usepackage{natbib}

\usepackage{graphicx}
\usepackage{tablefootnote}
\usepackage{longtable}
\usepackage{threeparttable}
\usepackage{wrapfig}
\usepackage{bm}
\usepackage{amssymb}

\title{Distilling Governing Laws and Source Input for Dynamical Systems from Videos}

\author{
Lele Luan$^1$
\and
Yang Liu$^2$
\and
Hao Sun$^{3}$\footnote{Corresponding Author}
\affiliations
$^1$Department of Civil and Environmental Engineering, Northeastern University, Boston, MA, USA\\
$^2$School of Engineering Sciences, University of the Chinese Academy of Sciences, Beijing, China\\
$^3$Gaoling School of Artificial Intelligence, Renmin University of China, Beijing, 100872, China
\emails
\emph{Emails}:~~~luan.l@northeastern.edu;~~~ liuyang22@ucas.ac.cn; ~~~haosun@ruc.edu.cn
}

\begin{document}

\maketitle

\begin{abstract}
Distilling interpretable physical laws from videos has led to expanded interest in the computer vision community recently thanks to the advances in deep learning, but still remains a great challenge. This paper introduces an end-to-end unsupervised deep learning framework to uncover the explicit governing equations of dynamics presented by moving object(s), based on recorded videos. Instead in the pixel (spatial) coordinate system of image space, the physical law is modeled in a regressed underlying physical coordinate system where the physical states follow potential explicit governing equations. A numerical integrator-based sparse regression module is designed and serves as a physical constraint to the autoencoder and coordinate system regression, and, in the meanwhile, uncover the parsimonious closed-form governing equations from the learned physical states. Experiments on simulated dynamical scenes show that the proposed method is able to distill closed-form governing equations and simultaneously identify unknown excitation input for several dynamical systems recorded by videos, which fills in the gap in literature where no existing methods are available and applicable for solving this type of problem.
\end{abstract}

\section{Introduction}

Discovery of governing equations (e.g., PDEs, ODEs) has the potential to advance our understanding, modeling and prediction of the behavior of complex dynamical systems. Increasing richness of collecting data and advances in machine learning gave rise to a new perspective of dynamical system modeling, e.g., data-driven discovery of governing equations recently \cite{brunton2016discovering}. Advances in deep learning, especially convolutional neural networks (CNNs), have led to keen interest in uncovering physical law directly from videos that record dynamical processes \cite{chen2021grounding}. In earlier deep learning approaches, explicit physical law was distilled from the motion trajectory which was extracted by supervised regression ahead \cite{de2018end}. Later, unsupervised moving object localization techniques promoted the parameter estimation of dynamical systems in an unsupervised scheme \cite{jaques2020physics}. While since the moving object is localized in image space, these approaches can only learn simple physical laws or their parameters which are modeled in pixel coordinate system such as billiards, gravity, spring, bouncing, etc. \cite{jaques2020physics,kossen2019structured}. Furthermore, in most of existing methods, the form of physical engine is given and incorporated with motion representative extraction module, which restricts their application to real scenarios where the mathematical form/structure of physical law is unknown.

\begin{figure*}[t!]
\centering
\includegraphics[width=0.71\textwidth]{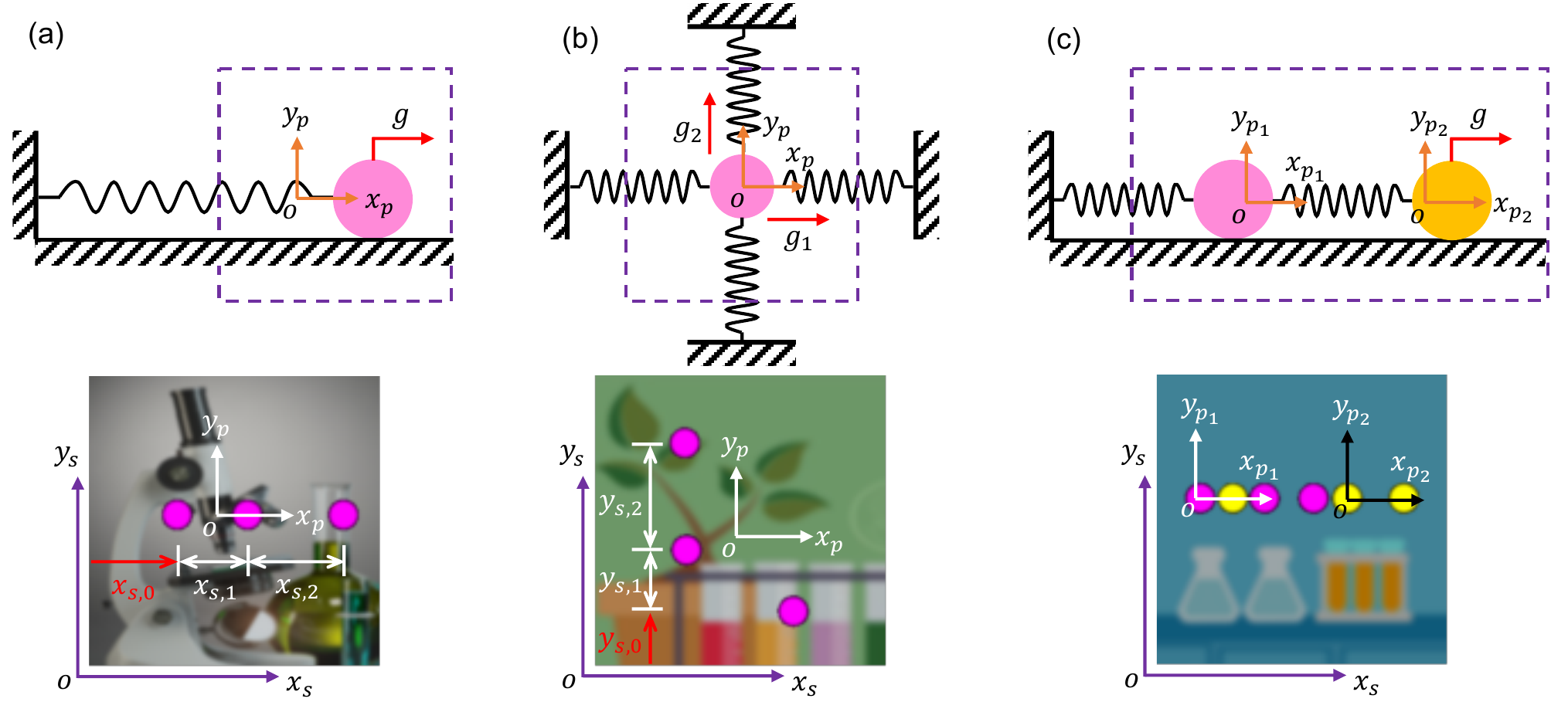}
\vspace{-6pt}
\caption{The studied dynamical systems excited by unknown inputs. The videos are generated by simulating the dynamical systems being recorded from the views of dash-line rectangular areas for governing equation discovery. (a) Single mass single degree of freedom system (SMSD); (b) Single mass two degrees of freedom system (SMTD); (c) Two masses two degrees of freedom system (TMTD). $g$, $g_1$ and $g_2$ represent the unified system inputs which will be identified along with the closed-form governing equation discovery. The second row gives the snapshots of generated videos showing the object(s) moving at different time steps.  $(x_s$--$o$--$y_s)$ represents the spatial (pixel) coordinate system which determines the positions of moving object(s) in image space while $(x_p$--$o$--$y_p)$ represents the regressed physical coordinate system to describe the physical states which present the underlying physical laws. Because there are two moving object(s) in TMTD system, two physical coordinate systems are regressed ($(x_{p_1}$--$o$--$y_{p_1})$ and $(x_{p_2}$--$o$--$y_{p_2})$) for their physical trajectories. In spatial coordinate system, the learned spatial coordinates may not be the real locations of the moving object(s) in image space (e.g., $x_{s,0}$ and $y_{s,0}$), while their relative positions (e.g., $x_{s,1}/x_{s,2}$ and $y_{s,1}/y_{s,2}$) can be hold by the spatial transformer (ST) based Coordinate-Consistent Decoder.}
\label{fig:dynamical_systems}
\vspace{-6pt}
\end{figure*}

Study on data-driven governing equations discovery started with and, today, still mainly focus on building mathematical models from given measurement of physical states (e.g., trajectory time series) \cite{kutz2016dynamic,brunton2016discovering,rudy2019data,udrescu2020ai}. Later, advances in deep learning led to expanded interests in physical law discovery from videos instead, a subset of physical scene understanding. Inspired by powerful feature extraction ability, deep neural network was initially leveraged to learn ``blur'' physics from videos, in which the physical laws are not expressed explicitly but simulated or condensed as physical modules \cite{fragkiadaki2016learning,ehrhardt2017learning}. Later, the incorporation of physical engines/modules with extracted object-based representations improved the description or prediction of physical videos \cite{purushwalkam2019bounce,wu2017learning,ye2018interpretable}. Although some of those approaches considered interpretable parameters in the physical modules like object mass, position, speed and friction \cite{wu2017learning,ye2018interpretable}, the physical laws are still not explicitly discovered and the physics are condensed as state space, or simulated as neural physics engines/modules.

In order to improve the interpretability of the discovered physical laws, learning explicit dynamics (e.g., closed-form governing equations or their parameters) has recently become more popular in physical scene understanding. Several hybrid methods take a data-driven approach to estimate real mechanical process from video sequences \cite{wu2015galileo,wu2016physics}, or model Newtonian physics via latent variable to predict motion trajectories in images \cite{mottaghi2016newtonian}. Since the discovery of explicit physical laws requires extracting the motion of moving object, two-step staggered discovery has become the most common strategy, where the physical law is distilled after the moving trajectory being extracted \cite{wu2017learning,de2018end,ehrhardt2018unsupervised}. Advances in unsupervised object localization techniques like based on spatial transformers (ST) \cite{kosiorek2018sequential,hsieh2018learning,ehrhardt2018unsupervised} and Position-Velocity Encoders (PVEs) \cite{jonschkowski2017pves} have enabled the explicit physical law discovery in an unsupervised scheme \cite{kossen2019structured,jaques2020physics}. However, these approaches require strong \textit{prior} knowledge on the structure of the physical law or governing equation (e.g., the equation form is given while the coefficients need to be discovered). Furthermore, for those methods, physics is modeled in pixel coordinates which restricts the discovery for complex dynamical systems (e.g., ODEs) where the physical states need to be described in another physical coordinate system.

Attempts also have been made to uncover the closed-form governing equations in the context of low-dimensional representations from videos or high-dimensional data. One approach to discover the governing equations and the associated coordinate system from high-dimensional data is proposed in \cite{champion2019data}, while this approach was only able to handle the ``video'' which can be explicitly expressed as a function of physical latent and fails to deal with real raw videos. The unsupervised physical parameter estimation method proposed in \cite{jaques2020physics} can only discover from videos the parameters of given physical laws which are established in the context of pixel coordinates. In addition, a two-step deep learning based method was developed for closed-form governing equation discovery from distorted videos \cite{udrescu2020symbolic}, while this method is not in an end-to-end scheme and cannot deal with systems excited by unknown input. Hence, uncovering governing equations straightly from raw videos still remains a grand challenge, especially when the source input is unknown. 

In this work, we propose an end-to-end unsupervised deep learning framework to uncover from videos the closed-form governing equations of dynamical systems subjected to unknown input. The task we intend to resolve, as shown in Figure \ref{fig:dynamical_systems}, demonstrates the paradigm we build seeking to simultaneously extract the physical states of moving object(s), uncover their governed closed-form equations, and identify the system input. Unlike the existing deep learning methods which typically discover physical laws from spatial/pixel coordinate trajectories of moving object(s), our method uncovers the explicit governing equations from the physical states in a regressed physical coordinate system instead, which makes it possible to discover more complex dynamical systems. Furthermore, we consider the form/structure of governing equation unknown \textit{a priori} and employ sparse regression to distill its formulation. In addition, the physical states are extracted not independently from the encoder-decoder and physical coordinate system regression but under the constraint of underlying physical law. The joint optimization not only helps the extraction of physical states, but also leads to the identification of closed-form governing equations and unknown input, which forms our key contribution.

\section{Methodology}

In this section, we first introduce the designed network architecture for governing equations discovery from videos, which is assembled with three segments: (1) an encoder-decoder to condense the high-dimensional image into low dimensional latent space, (2) physical coordinate system regression to create the mapping between the spatial/pixel coordinates of moving object(s) and their physical states for underlying explicit physical law, and (3) physical embedding which serves as a constraint for the extraction of physical states, and, meanwhile, uncovers their governed equation and the unknown excitation. The definition of loss function and the network training strategy are also provided in the following.

\subsection{Network Architecture}
\begin{figure*}[t!]
\centering
\includegraphics[width=0.84\textwidth]{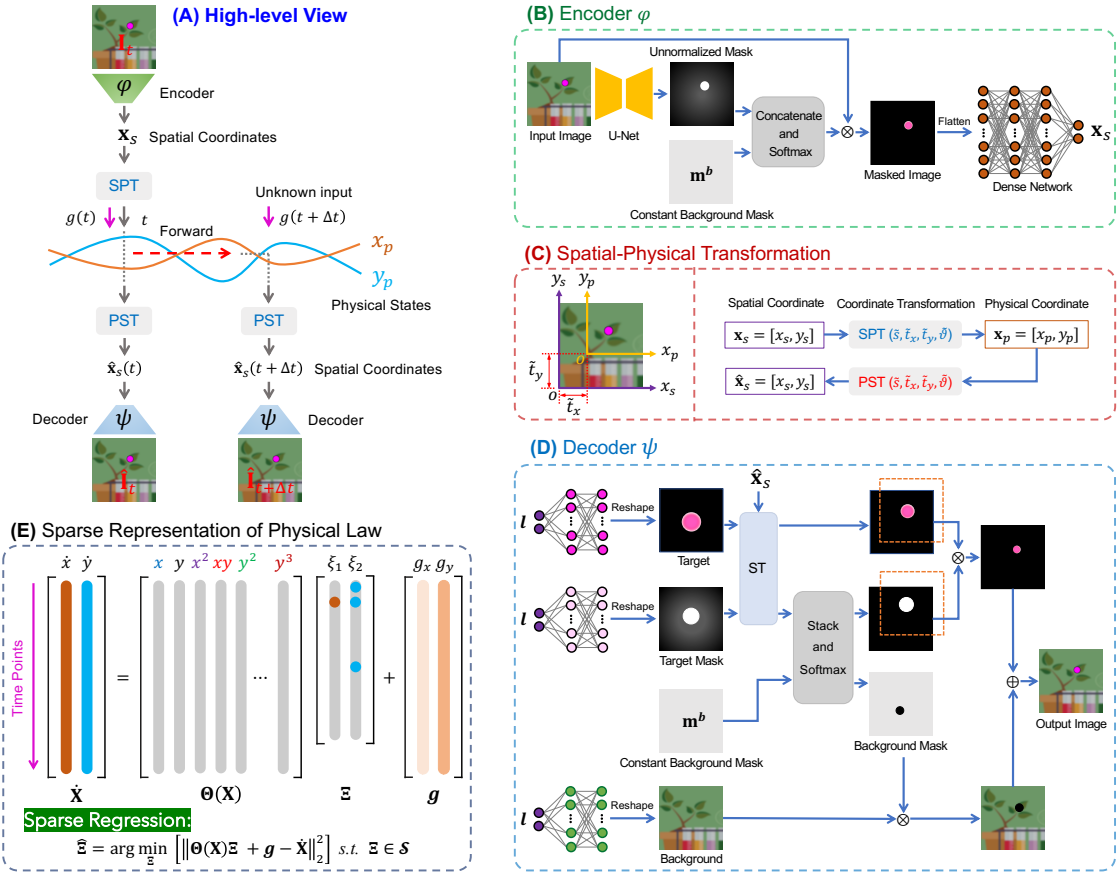}
\vspace{-6pt}
\caption{Schematic architecture of the proposed end-to-end unsupervised deep learning to simultaneously uncover the closed-form governing equation and identify the input of dynamical system (single moving object case) from videos. (A) represents high-level view of the network architecture involving the physics propagation. The unknown source input is considered in the forward propagation of physical state ($g(t)$ at $t$ and $g(t+\Delta t)$ at $t+\Delta t$). Single video frame $\mathbf{I}_t$ is fed into (B) Encoder $\varphi$ to capture the spatial coordinates of the moving object $\mathbf{x}_s$, which are mapped into physical states $\mathbf{x}_p$ by (C) Spatial-Physical Transformation (SPT) with factors of translation $\tilde{\mathbf{t}}=(\tilde{t}_x,\tilde{t}_y)^T$ and scaling $\tilde{s}$. The physical states then enter the Physical-Spatial Transformation (PST), with the same transformation factors as SPT, to reproduce the spatial coordinates for reconstructing the video frame with $\mathbf{\hat{I}}_t$ by (D) Coordinate-Consistent Decoder $\psi$. Because the governing equation form is unknown, the derivative of extracted physical states are expressed by a library of candidate functions $\mathbf{\Theta}(\mathbf{X})$ parameterized by unknown sparse coefficients $\mathbf{\Xi}\in\boldsymbol{\mathcal{S}}$ and the unknown input shown in (E). The physical state forward is also passed into PST to recover the spatial coordinates of moving object at forward snapshot and reconstruct this frame by decoder. In the decoder, the moving object and its masks, and the video background are learned by independent networks separately from a constant array $\bm{l}$. In Encoder and Decoder, $\mathbf{m}^b$ and $\bm{l}$ denote a tensor and an array with all elements of 1. Source codes for implementing this network are found in \url{https://github.com/LeleLuan/VideoDiscovery}.}
\label{fig:network_architecture}
\vspace{-6pt}
\end{figure*}

In order to learn the governing equations from videos, several components are considered in place. First of all, the high-dimensional image is condensed into low-dimensional latent variables, the spatial (pixel) coordinates of moving object(s), by the encoder-decoder. A new physical coordinate system is regressed to map the spatial coordinates into physical sates where the underlying physical law is presented. Since the potential governing equations are assumed to be parsimonious and can be described with fewest terms necessary, we leverage a library, composed of a finite number of pre-defined possible candidate terms in the context of physical states, to regress the equations parameterized by their unknown coefficients. Then the physical law constraint is imposed to the physical states by the sparse regression and the physics forward with a 4th-order Runge-Kutta method. In addition, the unknown input can be identified by embedding it into the physical constraint. The designed network architecture is depicted in Figure \ref{fig:network_architecture}. Figure \ref{fig:network_architecture}(A) shows the high-level view of the network architecture which involves the forward propagation of physics by considering two consecutive video frames. All components making up the high-level network are discussed as follows and more information about the network can be found in \textit{Supplementary Information (SI)}\footnote{see \url{https://github.com/LeleLuan/VideoDiscovery}} Section 2.1.

\paragraph{Encoder:}
A two-stage localization network \cite{jaques2020physics} is employed as encoder to condense the image space into low-dimensional latent variables. It takes a single video frame $\mathbf{I}_t$ as input and outputs a vector $\mathbf{x}_s=(x_{s1},y_{s1},x_{s2},y_{s2},...,x_{sn},y_{sn})$ corresponding to the 2D spatial coordinates of $n$ moving objects. First, the input frame is passed through a U-Net to produce unnormalized masks for moving object(s). The unnormalized masks are concatenated with a constant background mask $\mathbf{m}^b$ and then passed through a Softmax to produce the masks for the moving object(s) and background. The multiplication between input and masks are fed into a fully-connected network to obtain the spatial coordinates of the moving object(s). The output layer of the encoder depends on the number of coordinates needed to describe the object location. For a 2D physical system with image size of $H \times H$, the latent space has two variables and the activation of output layer has a saturating non-linearity $H/2\cdot\tanh{(\cdot)}+H/2$, which leads to the spatial coordinate of the moving object $\mathbf{x}_s$ with values in $[0, H]$.

\paragraph{Coordinate-Consistent Decoder:}
The decoder takes the spatial coordinates of moving object(s) $\mathbf{x}_s$ as input and outputs a reconstructed image. Instead of using conventional decoders, the decoder imposed by a fixed relationship between latent spatial coordinate and pixel-coordinate correspondence is employed to reconstruct the video frame. The Coordinate-Consistent Decoder developed in \cite{jaques2020physics} is built on the spatial transformer (ST) network. If the input spatial coordinate for one object to decoder is $\mathbf{x}_s=(x_s,y_s)$, the parameters $\omega$ of ST is to place the center of the writing attention window for moving object at this point. In this Coordinate-Consistent Decoder, the moving object(s) and their masks, and the background are learned by independent networks separately. Then the video frames are reconstructed by the composition of background and the moving object(s) the positions of which are determined by the input spatial coordinates with STs. It should be highlighted that, the Coordinate-Consistent Decoder may not able to capture the real (or absolute) locations of moving object(s) in image space (e.g., $x_{s,0}$ and $y_{s,0}$ in Figure \ref{fig:dynamical_systems}) especially for the videos with moving object(s) showing up in limited positions. While the relative locations of the object(s) moving at different time steps can be captured (e.g., $x_{s,1}/x_{s,2}$ and $y_{s,1}/y_{s,2}$ in Figure \ref{fig:dynamical_systems}), which plays a key role in the following physical trajectory extraction via Spatial-Physical Transformation.

\paragraph{Physical Coordinate System Regression:}
The Encoder and Coordinate-Consistent Decoder introduced above are expected to extract the spatial coordinates of all moving objects in the scene. An underlying physical coordinate system is then regressed to map the extracted spatial coordinates into physical states which present the potential governing equations. As shown in Figure \ref{fig:dynamical_systems}, if we assume the dynamics is present in a 2D space which is parallel to the image space, the transformation between spatial coordinates and physical 
states can be implemented by a standard 2D Cartesian coordinate transformation. For one moving object, with translation vector $\tilde{\mathbf{t}}=(\tilde{t}_x,\tilde{t}_y)^T$ and scaling factor $\tilde{s}$, the physical coordinate (state) $\mathbf{x}_p=(x_p,y_p)$ can be obtained from spatial coordinate by $\mathbf{x}^T_p=\mathcal{T}(\mathbf{x}^T_s)=\tilde{s}\big[\mathbf{x}^T_s-\tilde{\mathbf{t}}\big]$. Likewise, the physical to spatial transformation can be expressed as $\mathbf{x}^T_s = \tilde{\mathcal{T}}(\mathbf{x}^T_p)=1/\tilde{s}\mathbf{x}^T_p+\tilde{\mathbf{t}}$. Here, $\mathcal{T}$ and $\tilde{\mathcal{T}}$ share the same parameters of $\tilde{s}$ and $\tilde{\mathbf{t}}$ in a complete encoder-decoder. It should be noted that different coordinate systems are regressed for different moving objects when the scene has multiple objects. The transformation factors are trainable variables and will be learned in the whole network training. As shown in Figure \ref{fig:network_architecture}(A), with the Encoder $\varphi$, Coordinate-Consistent Decoder $\psi$ and coordinate transformers ($\mathcal{T}$ and $ \tilde{\mathcal{T}}$), a complete autoencoder is built for the condensation of video frames into latent space, the physical states of moving object(s). The standard autoencoder loss function can be expressed as: $\mathcal{L}_{recon} = \big\lVert\mathbf{I}_t-\psi(\tilde{\mathcal{T}}(\mathcal{T}(\varphi(\mathbf{I}_t))))\big\rVert_2^2$.

\paragraph{Physical Law Embedding and Discovery:}
For the studied dynamical systems shown in Figure \ref{fig:dynamical_systems}, the potential ODEs followed by moving object trajectories are distilled by sparse regression. When the dynamical system has one moving object with physical states $\mathbf{x}_p(t)=(x_p(t),y_p(t))^T$, the considered dynamics can be expressed by $\frac{d}{dt}\mathbf{x}_p(t)=\bm{f}(\mathbf{x}_p(t)) + \mathbf{g}(t)$, where $\mathbf{g}(t)$, a trainable array, represents the unknown system input which is constant for one dynamical system with different initial conditions. We seek a parsimonious model for $\bm{f}(\mathbf{x}_p(t))$, resulting in a function $\bm{f}$ that contains only a few active terms. If the derivatives of the physical states are calculated, the snapshots are stacked to form data matrices $\mathbf{X}=[\mathbf{x}_{p,1}, \mathbf{x}_{p,2}, \dotsc, \mathbf{x}_{p,m}]^T \in \mathbb{R}^{m\times 2}$ and $\dot{\mathbf{X}}=[\dot{\mathbf{x}}_{p,1}, \dot{\mathbf{x}}_{p,2}, \dotsc, \dot{\mathbf{x}}_{p,m}]^T \in \mathbb{R}^{m\times 2}$ where $m$ is the number of data points of the physical states. Although $\bm{f}$ is unknown, we can construct an extensive library of $n$ candidate functions $\mathbf{\Theta(\mathbf{\mathbf{X}})}=[\theta_1(\mathbf{X}), \theta_2(\mathbf{X}),  ..., \theta_n(\mathbf{X})] \in \mathbb{R}^{m \times n}$, where each $\mathbf{\theta}_j$ denotes a candidate term. The candidate function library is used to formulate an over-determined system $\dot{\mathbf{X}}=\mathbf{\Theta}(\mathbf{X})\mathbf{\Xi}$, where the unknown matrix $\mathbf{\Xi}=[\bm{\xi}_1, \bm{\xi}_2,...,\bm{\xi}_n]\in \mathbb{R}^{n \times 2}$ is the set of coefficients that determine the active terms from $\mathbf{\Theta(\mathbf{\mathbf{X}})}$ in the dynamics $\bm{f}$. For the candidate function library, $\theta_i(\mathbf{X})$ can be any candidate function that may describe the system dynamics $\bm{f}(\mathbf{x}(t))$ such as $\theta_i(\mathbf{X})=\mathbf{X}^2$. By solving the optimization, the model of system dynamics can be identified $\frac{d}{dt}\mathbf{x}_p(t) \approx \mathbf{\Theta}(\mathbf{x}_p(t))\mathbf{\Xi} + \mathbf{g}(t)$. The coefficients $\mathbf{\Xi}$ are learned concurrently with the neural network parameters as part of the training process. With the time derivative of physical states $\mathbf{x}_p(t)$ being calculated by the central difference method, one can enforce accurate modeling of the dynamics by incorporating the following physical state derivative term into the loss function: $\mathcal{L}_{\dot{\mathbf{x}}_p} = \big\lVert\frac{d(\mathcal{T}(\varphi(\mathbf{I}_t))}{dt} - \mathbf{\Theta}((\mathcal{T}(\varphi(\mathbf{I}_t)))^T)\mathbf{\Xi}+\mathbf{g}(t)\big\rVert_2^2$.

As shown in Figure \ref{fig:network_architecture}, the forward propagation of physics also brings an additional physical constraint which is imposed to the physical states. With the true vector field $f(\mathbf{x}_p(t))$ being estimated by $\mathbf{\Theta}(\mathbf{x}_p(t))\mathbf{\Xi}$, integrating over a segment of time $t_j$ to $t_{j+1}$ gives the integrated vector field, or flow map $\mathbf{x}_p(j+1))=\mathbf{x}_p(j)+\int_{t_j}^{t_{j+1}}(\mathbf{\Theta}(\mathbf{x}_p(\tau))\mathbf{\Xi}+\mathbf{g}(\tau))d\tau$. A 4th-order Runge-Kutta (RK4) numerical method is employed to simulate the dynamics 1-step forward as $\mathbf{x}_p(j+1)=\mathrm{RK4}(\mathbf{x}_p(j),\mathbf{g}(j),\mathbf{g}(j+1))$. Then the forward video frame can be reconstructed via the decoder as $\hat{\mathbf{I}}_{j+1}=\psi(\tilde{\mathcal{T}}(\mathbf{x}_p(j+1)))$, which leads to the forward video frame reconstruction loss: $\mathcal{L}_{int} = \big\lVert{\mathbf{I}}_{j+1}-\psi\big(\tilde{\mathcal{T}}\big(\mathbf{x}_p(j+1)\big)\big)\big\rVert_2^2$.

\subsection{Loss Functions and Network Training}
In addition to three loss terms given above, an $\ell_{0.5}$ regularizer $\mathcal{L}_{reg}$ on the sparse regression coefficients $\mathbf{\Xi}$ is included, which promotes sparsity of the coefficients and therefore encourages a parsimonious model for the dynamics. The combination of 4 loss terms leads to the overall loss function:
\begin{equation}
\label{loss}
\mathcal{L}_{total}=\mathcal{L}_{recon}+\lambda_1\mathcal{L}_{\dot{\mathbf{x}}_p}+\lambda_2\mathcal{L}_{int}+\lambda_3\mathcal{L}_{reg}
\end{equation}
where the hyperparameters $\lambda_1$, $\lambda_2$, $\lambda_3$ determine the relative weighting of the 3 loss function terms. The detailed definition of loss functions is given in \textit{SI} Section 2.2. The network is trained using a multi-step training strategy to succeed the discovery, which includes $(\mathbf{\romannumeral1})$ pre-training, $(\mathbf{\romannumeral2})$ total loss training, $(\mathbf{\romannumeral3})$ sequential thresholding, and $(\mathbf{\romannumeral4})$ refinement training. In the designed network, the trainable variables include variables in encoder-decoder, transformation factors $(\tilde{t}_x,\tilde{t}_y)^T$ and $\tilde{s}$ in physical coordinate system regression, coefficients for candidate functions $\mathbf{\Xi}$ in sparse regression, and system input array $\mathbf{g}$. Although some of the trainable variables are not independent, e.g., $\mathbf{\Xi}$ and $\mathbf{g}$, the network is still able to achieve the discovery thanks to the constraint of physical law and the sparsity of governing equation. More information about the multi-step training strategy and the choice of hyperparameters are provided in \textit{SI} Section 2.3 and \textit{SI} Section 2.4.

\section{Experiments}

We demonstrate the efficacy of our proposed method on several dynamical systems shown in Figure \ref{fig:dynamical_systems} (including SMSD, SMTD, TMTD). The studied videos are generated by plotting the lumped mass(es) according to the simulated physical trajectories on different backgrounds. More details of the dataset generation are given in \textit{SI} Section 1. Our method aims to uncover the closed-form governing equations for the moving object(s) and identify the unknown source input. It should be noted that, since the studied dynamical systems follow second-order ODEs, the discovery is conducted by transferring the underlying equations into first-order state-space models. More information about the discovery on state-space modeling is given in \textit{SI} Section 2.5. In addition, the robustness of this method on discovery to noise, ablation studies, and baseline comparison are also discussed in this section.

\subsection{Discovery Result}
\begin{figure}[t!]
\centering
\includegraphics[width=0.482\textwidth]{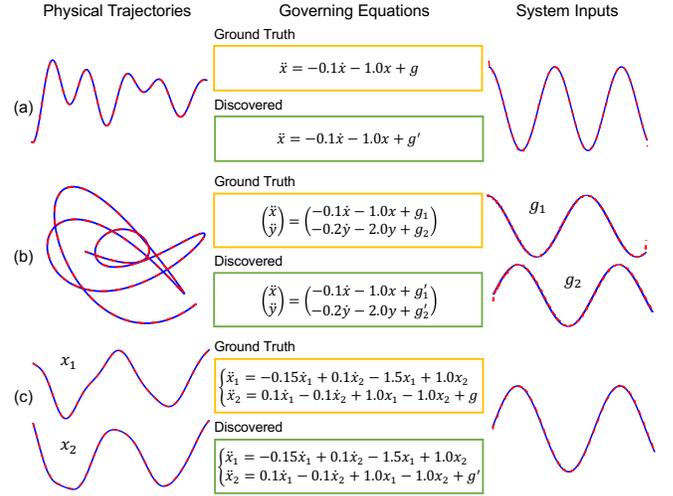}
\vspace{-12pt}
\caption{Discovery results of the studied dynamical systems, (a) SMSD, (b) SMTD and (c) TMTD. The ground truth physical trajectories and inputs are presented by solid lines, while the learned are presented by dash lines. The physical trajectories and inputs shown are the results after scaling and translation. In the SMTD system, $g_1$ and $g_2$ denote the inputs applied to horizontal and vertical direction of the mass. In the TMTD system, $x_1$ and $x_2$ denotes the physical states of left and right moving object, respectively.}
\label{fig:discovery}
\vspace{0pt}
\end{figure}

The discovery results for the studied dynamical systems are presented in Figure \ref{fig:discovery}, where the physical trajectories, their governing equations, and the external excitation are uncovered. It shows that the governing equations especially their coefficients are identified exactly the same as the ground truth. It should be noted that, in the discovered ODEs, both candidate function terms $\mathbf{\Theta}(\mathbf{X})$ and input $\mathbf{g}$ are unknown. The discovered equations are still satisfied although physical states $\mathbf{x}_p$ and input $\mathbf{g}$ have scaling or drift discrepancy to the ground truth. While after scaling and translation, the physical states and inputs learned from the network match to the ground truth very well. The discovery result for the TMTD system also shows that our method is able to handle the scenarios when multiple moving objects show up in the scene. Because different moving objects may follow different physical laws in different physical coordinate systems, the network is equipped with multiple coordinate regression and sparse regression for both physical trajectory extraction and governing equation discovery. Furthermore, because the dynamics presented by those two moving objects are not independent, the physical states ($x_1$, $x_2$, $y_1$ and $y_2$) and their derivatives ($\dot{x}_1$, $\dot{x}_2$, $\dot{y}_1$ and $\dot{y}_2$) are considered in building the sparse regression for their governing equation discovery. In addition to the studied linear dynamical system, our method was also tested on discovering the governing equation of a nonlinear dynamical system (see \textit{SI} Section 4.4). The detailed information for each dynamical system discovery is given in \textit{SI} Section 4.

\begin{figure}[t!]
\centering
\includegraphics[width=0.483\textwidth]{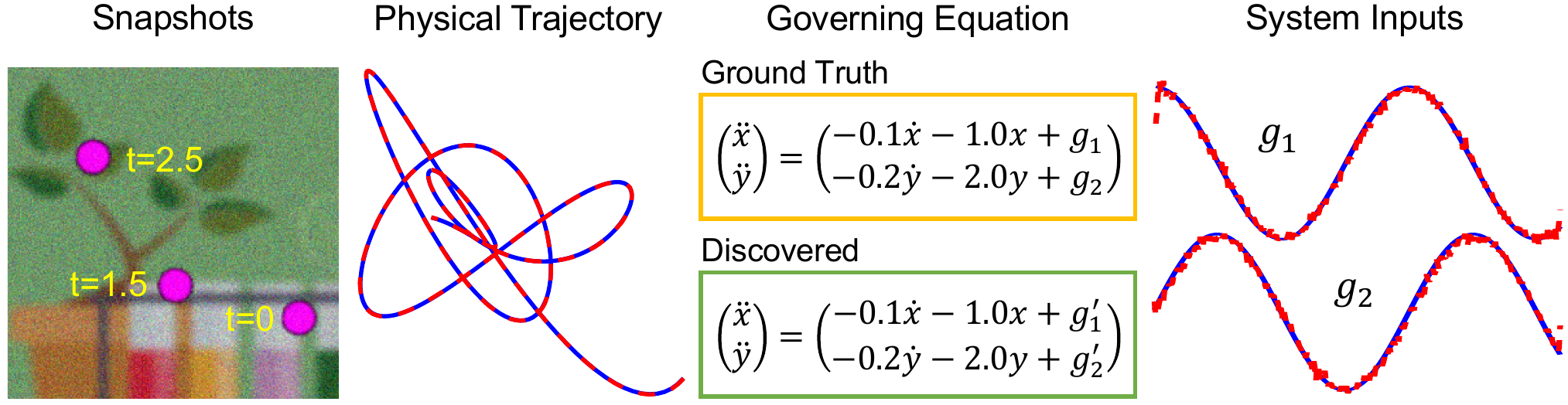}
\vspace{-12pt}
\caption{Discovery result for SMTD system from noisy videos.}
\label{fig:discovery_noise}
\vspace{-6pt}
\end{figure}

\subsection{Robustness to Noise}
The robustness of our proposed method to noise is further tested by discovering the governing equation from videos with noise. Here, we take the SMTD system as the testing example where Gaussian noise with zero mean and variance $2\times10^{-5}$ is added to all video frames. The discovery result is presented in Figure \ref{fig:discovery_noise}. It shows that, although the noise in the videos is against the assumption that both moving object(s) and background are constant for all images in the ST-based Coordinate-Consistent Decoder, our method still achieves the correct discovery. Due to the effect of noise, compared to the discovery from videos without noise, the identified system input is noisier. Nevertheless, the governing equations and the physical trajectory are still uncovered and extracted correctly.

\subsection{Ablation Study}
In the proposed method, the spatial coordinates of moving object(s) learned from Encoder and Coordinate-Consistent Decoder are mapped into physical states by a regressed physical coordinate system. Here, the ablation study is implemented by removing the Spatial-Physical Transformation and then imposing the physical law constraint to spatial coordinates directly to test the necessity of physical coordinate system regression. The SMSD system is taken as the example for this ablation study. In the network training, the physical states derivative loss $\mathcal{L}_{\dot{\mathbf{x}}_p}$ is quite large after pre-training and this loss term does not converge at all in optimizing the pre-trained model with total loss function. This is because the underlying physical law is not presented in pixel coordinate, adding physical law constraint to spatial coordinate directly will lead to high physical derivative loss and this loss term does not converge in the total loss optimization.

\subsection{Baseline Comparison}
\begin{figure}[t!]
\centering
\includegraphics[width=0.45\textwidth]{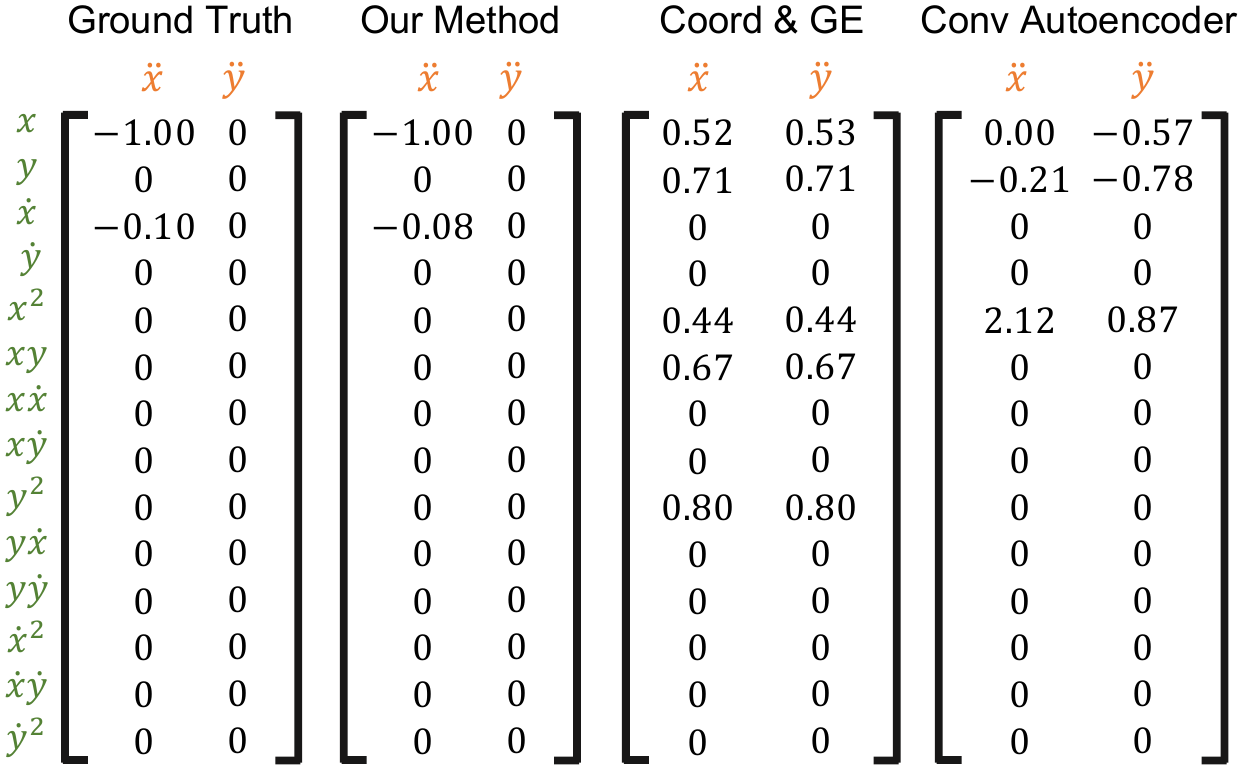}
\vspace{-6pt}
\caption{Discovery results for the baselines. The coefficient matrices given here are the learned candidate function coefficients after sequential thresholding in training process from different methods. Coord \& GE represents the coordinate and governing equation discovery method developed in \cite{champion2019data}. Conv autoencoder represents the designed baseline in which the video image is condensed by conventional convolutional autoencoder.}
\label{fig:baseline}
\vspace{-6pt}
\end{figure}

We found that the literature remains scant in uncovering governing equations for dynamical systems with unknown input directly from videos. The most related work was presented in \cite{champion2019data} where an autoencoder-based deep learning model was built to discover low-dimensional dynamics and their associated coordinates from high-dimensional data. As a baseline, this coordinate and governing equation discovery method must be adapted slightly to account for unknown source input. In this baseline, the SINDy module has the same candidate functions as the sparse regression used in our discovery. Besides, the original RGB videos are converted into gray scale to calculate the required derivatives. More details on this baseline are provided in \textit{SI} Section 5. The SMSD system is taken as an example. The training process shows that the reconstruction loss and SINDy loss in $\ddot{\mathbf{z}}$ ($\dot{\mathbf{z}}$ for the first-order ODE discovery) can achieve small values (e.g., $2\times 10^{-4}$ and $4\times 10^{-7}$ respectively) while the SINDy loss in $\ddot{\mathbf{x}}$ remains almost constant (e.g., 0.34). After sequential thresholding in the training process, the identified coefficient matrix as shown in Figure \ref{fig:baseline} has no sparsity. The failure of this baseline method is because the raw video images cannot be treated as an explicit function of some latent variables, the case for which this baseline method is only applicable.

In addition, we replace the coordinate-consistent encoder-decoder shown in Figure \ref{fig:network_architecture} by a conventional convolutional encoder-decoder and take the resulting method as another baseline. In this baseline, the video frame is condensed into low-dimensional latent variables by convolutional layers and the Spatial-Physical Transformation module is removed. The physical law constraint is still imposed to the latent variables. More information about this convolutional encoder-decoder baseline is given in \textit{SI} Section 5. The identified coefficient matrix is shown in Figure \ref{fig:baseline}. Once the model is trained, the autoencoder loss, physical states derivative loss and forward frame reconstruction loss become very small (e.g., $2.0 \times 10^{-5}$, $0.0020$ and $2.0 \times 10^{-5}$), which demonstrates that this network is able to distill a physical law from videos. However, since the extracted latent variables cannot correctly represent the location-based physical states, the method fails to uncover the underlying physical law. Furthermore, the fixed relationship between the physical states and real locations of the moving object(s) cannot be guaranteed by the conventional autoencoder.

\section{Conclusions}

This paper presents an end-to-end unsupervised deep learning scheme to uncover the explicit interpretable physical laws from raw videos that record moving object(s) representing dynamical systems excited by unknown input. Our approach takes the advantage of the spatial transformer (ST) based Coordinate-Consistent Decoder to capture the relative location of object(s) moving at different snapshots in the image space. In the physical coordinate system regression, the underlying coordinate system(s) are learned to map the spatial/pixel coordinates of moving object(s) into physical states where the dynamics may be explicitly and parsimoniously expressed. The form of governing equations for the extracted physical states is modeled by parametric linear combination of candidate functions, leading to a sparse regression problem. The sparse regression not only enables the physical law to be embedded into the network, but also provides a physical constraint to the extraction of physical states from the autoencoder and coordinate system regression. The efficacy of our proposed method is validated by uncovering the governing equations of several dynamical systems with different numbers of object in the scene. The robustness of our proposed method to noise is also tested. Our work is the first attempt to discover the interpretable physical laws from raw videos for dynamical systems with unknown input excitation. Our method also holds some limitations,. For example, it cannot deal with non-stationary background, video with warp, and moving objects in a 3D space. We are motivated to address these challenges in our ongoing and future study.

\section*{Acknowledgments}
The work is supported by the Beijing Outstanding Young Scientist Program (No. BJJWZYJH012019100020098) as well as the Intelligent Social Governance Platform, Major Innovation \& Planning Interdisciplinary Platform for the ``Double-First Class'' Initiative, Renmin University of China. Source codes and datasets are found in \url{https://github.com/LeleLuan/VideoDiscovery}. 

\bibliographystyle{named}
\bibliography{ijcai22}

\begin{thebibliography}{}

\bibitem[\protect\citeauthoryear{Belbute-Peres \bgroup \em et al.\egroup
  }{2018}]{de2018end}
Filipe de~A Belbute-Peres, Kevin~A Smith, Kelsey~R Allen, Joshua~B Tenenbaum,
  and J~Zico Kolter.
\newblock End-to-end differentiable physics for learning and control.
\newblock {\em Advances in Neural Information Processing Systems},
  31:7178--7189, 2018.

\bibitem[\protect\citeauthoryear{Brunton \bgroup \em et al.\egroup
  }{2016}]{brunton2016discovering}
Steven~L Brunton, Joshua~L Proctor, and J~Nathan Kutz.
\newblock Discovering governing equations from data by sparse identification of
  nonlinear dynamical systems.
\newblock {\em Proceedings of the National Academy of Sciences},
  113(15):3932--3937, 2016.

\bibitem[\protect\citeauthoryear{Champion \bgroup \em et al.\egroup
  }{2019}]{champion2019data}
Kathleen Champion, Bethany Lusch, J~Nathan Kutz, and Steven~L Brunton.
\newblock Data-driven discovery of coordinates and governing equations.
\newblock {\em Proceedings of the National Academy of Sciences},
  116(45):22445--22451, 2019.

\bibitem[\protect\citeauthoryear{Chen \bgroup \em et al.\egroup
  }{2021}]{chen2021grounding}
Zhenfang Chen, Jiayuan Mao, Jiajun Wu, Kwan-Yee~Kenneth Wong, Joshua~B
  Tenenbaum, and Chuang Gan.
\newblock Grounding physical concepts of objects and events through dynamic
  visual reasoning.
\newblock {\em International Conference on Learning Representations}, 2021.

\bibitem[\protect\citeauthoryear{Ehrhardt \bgroup \em et al.\egroup
  }{2017}]{ehrhardt2017learning}
Sebastien Ehrhardt, Aron Monszpart, Niloy~J Mitra, and Andrea Vedaldi.
\newblock Learning a physical long-term predictor.
\newblock {\em arXiv preprint arXiv:1703.00247}, 2017.

\bibitem[\protect\citeauthoryear{Ehrhardt \bgroup \em et al.\egroup
  }{2018}]{ehrhardt2018unsupervised}
Sebastien Ehrhardt, Aron Monszpart, Niloy Mitra, and Andrea Vedaldi.
\newblock Unsupervised intuitive physics from visual observations.
\newblock In {\em Asian Conference on Computer Vision}, pages 700--716.
  Springer, 2018.

\bibitem[\protect\citeauthoryear{Fragkiadaki \bgroup \em et al.\egroup
  }{2016}]{fragkiadaki2016learning}
Katerina Fragkiadaki, Pulkit Agrawal, Sergey Levine, and Jitendra Malik.
\newblock Learning visual predictive models of physics for playing billiards.
\newblock In {\em International Conference on Learning Representations}, 2016.

\bibitem[\protect\citeauthoryear{Hsieh \bgroup \em et al.\egroup
  }{2018}]{hsieh2018learning}
Jun-Ting Hsieh, Bingbin Liu, De-An Huang, Li~Fei-Fei, and Juan~Carlos Niebles.
\newblock Learning to decompose and disentangle representations for video
  prediction.
\newblock In {\em Advances in Neural Information Processing Systems}, pages
  515--524, 2018.

\bibitem[\protect\citeauthoryear{Jaques \bgroup \em et al.\egroup
  }{2020}]{jaques2020physics}
Miguel Jaques, Michael Burke, and Timothy Hospedales.
\newblock Physics-as-inverse-graphics: Unsupervised physical parameter
  estimation from video.
\newblock In {\em International Conference on Learning Representations}, 2020.

\bibitem[\protect\citeauthoryear{Jonschkowski \bgroup \em et al.\egroup
  }{2017}]{jonschkowski2017pves}
Rico Jonschkowski, Roland Hafner, Jonathan Scholz, and Martin Riedmiller.
\newblock Pves: Position-velocity encoders for unsupervised learning of
  structured state representations.
\newblock {\em arXiv preprint arXiv:1705.09805}, 2017.

\bibitem[\protect\citeauthoryear{Kosiorek \bgroup \em et al.\egroup
  }{2018}]{kosiorek2018sequential}
Adam~R Kosiorek, Hyunjik Kim, Ingmar Posner, and Yee~Whye Teh.
\newblock Sequential attend, infer, repeat: generative modelling of moving
  objects.
\newblock In {\em Advances in Neural Information Processing Systems}, pages
  8615--8625, 2018.

\bibitem[\protect\citeauthoryear{Kossen \bgroup \em et al.\egroup
  }{2020}]{kossen2019structured}
Jannik Kossen, Karl Stelzner, Marcel Hussing, Claas Voelcker, and Kristian
  Kersting.
\newblock Structured object-aware physics prediction for video modeling and
  planning.
\newblock {\em International Conference on Learning Representations}, 2020.

\bibitem[\protect\citeauthoryear{Kutz \bgroup \em et al.\egroup
  }{2016}]{kutz2016dynamic}
J~Nathan Kutz, Steven~L Brunton, Bingni~W Brunton, and Joshua~L Proctor.
\newblock {\em Dynamic mode decomposition: data-driven modeling of complex
  systems}.
\newblock SIAM, 2016.

\bibitem[\protect\citeauthoryear{Mottaghi \bgroup \em et al.\egroup
  }{2016}]{mottaghi2016newtonian}
Roozbeh Mottaghi, Hessam Bagherinezhad, Mohammad Rastegari, and Ali Farhadi.
\newblock Newtonian scene understanding: Unfolding the dynamics of objects in
  static images.
\newblock In {\em Proceedings of the IEEE Conference on Computer Vision and
  Pattern Recognition}, pages 3521--3529, 2016.

\bibitem[\protect\citeauthoryear{Purushwalkam \bgroup \em et al.\egroup
  }{2019}]{purushwalkam2019bounce}
Senthil Purushwalkam, Abhinav Gupta, Danny~M Kaufman, and Bryan Russell.
\newblock Bounce and learn: Modeling scene dynamics with real-world bounces.
\newblock {\em International Conference on Learning Representations}, 2019.

\bibitem[\protect\citeauthoryear{Rudy \bgroup \em et al.\egroup
  }{2019}]{rudy2019data}
Samuel Rudy, Alessandro Alla, Steven~L Brunton, and J~Nathan Kutz.
\newblock Data-driven identification of parametric partial differential
  equations.
\newblock {\em SIAM Journal on Applied Dynamical Systems}, 18(2):643--660,
  2019.

\bibitem[\protect\citeauthoryear{Udrescu and Tegmark}{2020}]{udrescu2020ai}
Silviu-Marian Udrescu and Max Tegmark.
\newblock {AI Feynman}: A physics-inspired method for symbolic regression.
\newblock {\em Science Advances}, 6(16):2631, 2020.

\bibitem[\protect\citeauthoryear{Udrescu and
  Tegmark}{2021}]{udrescu2020symbolic}
Silviu-Marian Udrescu and Max Tegmark.
\newblock Symbolic pregression: discovering physical laws from distorted video.
\newblock {\em Physical Review E}, 103(4):043307, 2021.

\bibitem[\protect\citeauthoryear{Wu \bgroup \em et al.\egroup
  }{2015}]{wu2015galileo}
Jiajun Wu, Ilker Yildirim, Joseph~J Lim, Bill Freeman, and Josh Tenenbaum.
\newblock Galileo: Perceiving physical object properties by integrating a
  physics engine with deep learning.
\newblock {\em Advances in Neural Information Processing Systems}, 28:127--135,
  2015.

\bibitem[\protect\citeauthoryear{Wu \bgroup \em et al.\egroup
  }{2016}]{wu2016physics}
Jiajun Wu, Joseph~J Lim, Hongyi Zhang, Joshua~B Tenenbaum, and William~T
  Freeman.
\newblock Physics 101: Learning physical object properties from unlabeled
  videos.
\newblock In {\em BMVC}, volume~2, page~7, 2016.

\bibitem[\protect\citeauthoryear{Wu \bgroup \em et al.\egroup
  }{2017}]{wu2017learning}
Jiajun Wu, Erika Lu, Pushmeet Kohli, Bill Freeman, and Josh Tenenbaum.
\newblock Learning to see physics via visual de-animation.
\newblock In {\em Advances in Neural Information Processing Systems}, pages
  153--164, 2017.

\bibitem[\protect\citeauthoryear{Ye \bgroup \em et al.\egroup
  }{2018}]{ye2018interpretable}
Tian Ye, Xiaolong Wang, James Davidson, and Abhinav Gupta.
\newblock Interpretable intuitive physics model.
\newblock In {\em Proceedings of the European Conference on Computer Vision
  (ECCV)}, pages 87--102, 2018.

\end{thebibliography}

\end{document}